\documentclass[11pt, letterpaper]{article}

\usepackage[margin=1.2in]{geometry} 
\usepackage{mathptmx} 
\usepackage[utf8]{inputenc}
\usepackage[T1]{fontenc}
\usepackage{authblk} 

\usepackage{amsmath, amssymb, amsthm} 
\usepackage{algorithmic, algorithm}
\usepackage{listings}
\usepackage{xcolor}
\usepackage{graphicx}
\usepackage{float}
\usepackage{cite}
\usepackage[hidelinks]{hyperref}
\usepackage{booktabs}
\usepackage{caption} 
\lstdefinestyle{mystyle}{
    backgroundcolor=\color{lightgray!20}, 
    basicstyle=\ttfamily\small,
    keywordstyle=\color{blue},
    commentstyle=\color{green!40!black},
    stringstyle=\color{red},
    breaklines=true,
    frame=single
}
\usepackage{authblk}

\title{\textbf{Persistent Homology–Guided Frequency Filtering for Image Compression}}

\author[1,2]{Anil Chintapalli}
\author[1]{Peter Tenholder}
\author[1]{Henry Chen}
\author[1,3]{Arjun Rao}

\affil[1]{North Carolina School of Science and Mathematics, Morganton, NC}
\affil[2]{University of North Carolina at Chapel Hill, Chapel Hill, NC}
\affil[3]{University of Texas at Austin, Austin, TX}

\date{December 2025} 
\begin{document}

\maketitle

\begin{abstract}
\noindent \textbf{Abstract.} Feature extraction in noisy image datasets presents many challenges in model reliability. In this paper, we use the discrete Fourier transform in conjunction with persistent homology analysis to extract specific frequencies that correspond with certain topological features of an image. This method allows the image to be compressed and reformed while ensuring that meaningful data can be differentiated. Our experimental results show a level of compression comparable to that of using JPEG using six different metrics. The end goal of persistent homology-guided frequency filtration is its potential to improve performance in binary classification tasks (when augmenting a Convolutional Neural Network) compared to traditional feature extraction and compression methods. These findings highlight a useful end result: enhancing the reliability of image compression under noisy conditions.
\end{abstract}

\vspace{1em}

\section{Introduction}
Feature extraction plays a crucial role in computer vision, providing meaningful representations of raw image data for machine learning tasks such as classification, segmentation, and object recognition. Contextually, variations in pixel values can distort meaningful structures due to sensor limitations, environmental conditions, or data compression \cite{Algan_2021, callet2024dft}. Persistent homology analyzes how the topological features of data persist as the data are filtered. As a result, homological methods identify features that are robust to small changes in the data, measuring their stability across varying scales. Traditional feature extraction methods, such as edge detection, texture analysis, and deep learning-based techniques, can be highly sensitive to these variations, leading to poor model performance and inaccurate predictions. In order to address these problems introduced by traditional feature extraction, we seek to use persistent homology (PH) and the discrete Fourier transform (DFT) to better represent the frequency and significance of topological features of an image \cite{franzenclassification}.

Motivation for this project came from the 2020 paper by Dong, Hu, Zhou and Lin \cite{Dong2020-vr} who used persistent homology as a means to classify patterns within porous structures. Their research aimed to test different ways to vectorize persistent homology data, creating a new representation of porous structures that could be used for a machine learning model. We were also inspired by
Victoria Callet's work \textit{DFT and Persistent Homology for Topological Musical
Data Analysis} which uses a novel idea of combining the DFT and persistent homology to better classify musical structures \cite{callet2024dft}. In this approach, short bars of music are transformed using the DFT and then represented as a point cloud. This is derived from their Fourier coefficients and then analyzed with persistent homology, creating a better understanding of the topology of music.

Our paper aims to build upon previous work by combining the discrete Fourier transform and persistent homology to perform image processing tasks, including compression and denoising. This can be further extended to improve image classification in noisy data without augmentation.

\section{Mathematical Background}

\subsection{Persistent Homology} \label{PH}

Persistent Homology was formally introduced in the 2002 paper by Edelsbrunner, Letscher, and Zomorodian \cite{edelsbrunner_topological_2002} as a way to better classify topological changes as a feature or noise. This new methodology was aimed to topologically simplify computer graphics and modeling. To formalize these ideas, we first introduce the fundamental building blocks of PH: simplices and complexes.

\begin{figure}[h]
\centering
\includegraphics[width=0.9\textwidth]{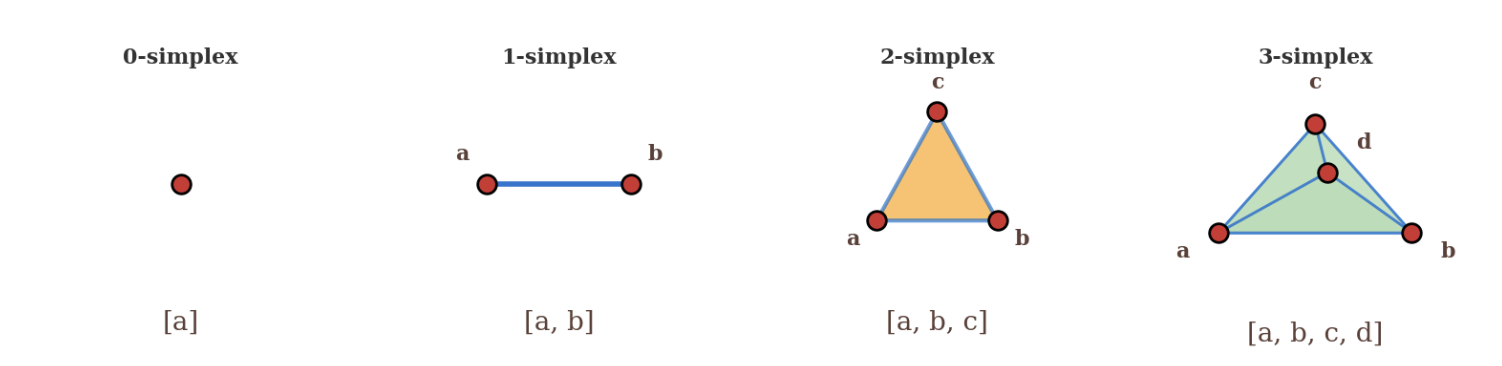}
\caption{Oriented \(k\)-simplices with \(k = 0, 1, 2, 3\). Adapted from \cite{edelsbrunner2010computational}.}
\label{simplical_complex}
\end{figure}

 A \emph{simplex} is equal to the join between $n$ points, which is a complex $n-1$ dimensional polyhedron. If the points $n$ that make up the simplex are the $n$ standard basis vectors for $\mathbb{R}^n$, then the join of these points is the simplex  $\Delta^{n-1} = \{(t_1, t_2, \cdots , t_n)\in \mathbb{R}^n \ | \  t_1 + t_2 + \cdots + t_n = 1 \  \text{and} \  t_i \ge 0\}$ \cite{hatcher_algebraic_2002}. (see Figure~\ref{simplical_complex} for examples of simplices)

A \emph{simplicial complex} \(K\), is a union of simplices such that every face of a simplex in \(K\) is also in \(K\), and the intersection of any two simplices in \(K\) is either empty or a shared face.

In particular, a \emph{k-dimensional complex} is a simplicial complex with 2 requirements:
\begin{enumerate}
    \item At least one k-simplex
    \item No simplices of dimension strictly greater than k
\end{enumerate}
To analyze topological features across different spatial scales, we build a \emph{filtration}, a nested sequence of simplicial complexes \( K_0 \subseteq K_1 \subseteq \cdots \subseteq K_n \), where each \( K_i \) includes more simplices than the last.

The homology of a complex measures and classifies holes (or cycles) within a topological space. We assign Betti numbers to count the number of $n$ dimensional holes in a complex. 

The $n$th Betti number represents the rank of $n$th homology group $H_n$, which is a vector space over $\mathbb{Z}$ generated by cycles. For example:
\begin{enumerate}
    \item $H_0$ - Holes generated by connected components
    \item $H_1$ - Holes generated by independent "loops/1-cycles" in a space
    \item $H_2$ - Holes generated by surfaces, like a sphere
\end{enumerate}

\begin{figure}[h]
  \centering
  \includegraphics[width=0.8\textwidth]{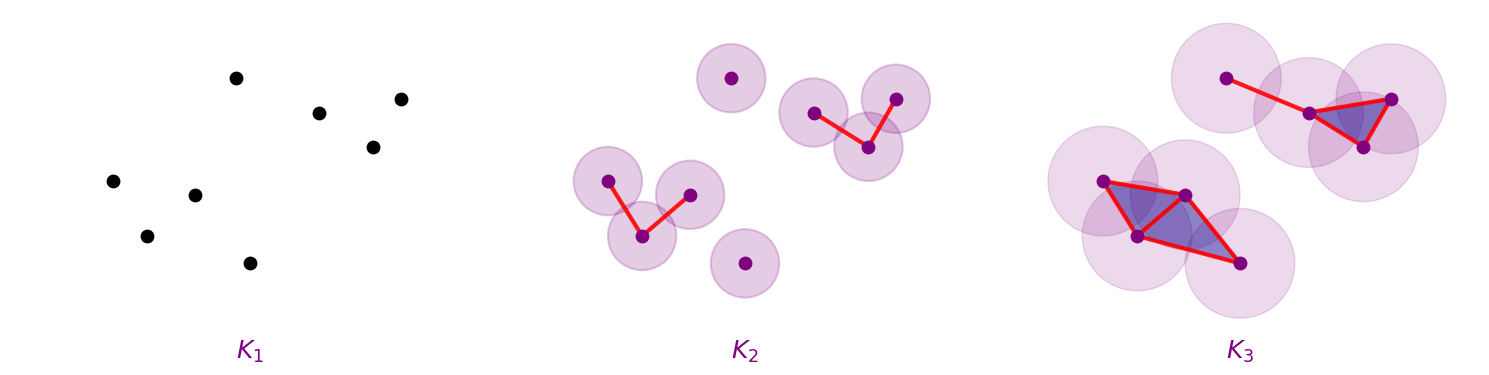}
    \caption{Diagram of the filtration of simplicial complexes used to compute the persistent homology of a point cloud.}.
  \label{fig:ph}
\end{figure}

With these foundations, we can compute persistent homology data from a given point cloud. In order to extract data, we gradually increase the radius of circles around each of the points. As the radii of the circles grow, the complex changes as points become connected to other points around them, causing the homology of the resulting complex to evolve continuously (see Figure~\ref{fig:ph}). When trying to extract data, we look at how holes in multiple dimensions within these simplicial complexes are created or destroyed. The creation of a hole (formally a cycle) is called its birth and the destruction of a hole called its death. We can mark birth and death times of the holes as our data.

For a given homological feature $i$, let $b_i$ indicate its birth time and $d_i$ indicate its death time. The persistence of this $ith$ feature is defined as follows:
\begin{center}
    $pers_i = d_i - b_i$
\end{center}

\begin{figure}[h]
  \centering
  \includegraphics[width=1\textwidth]{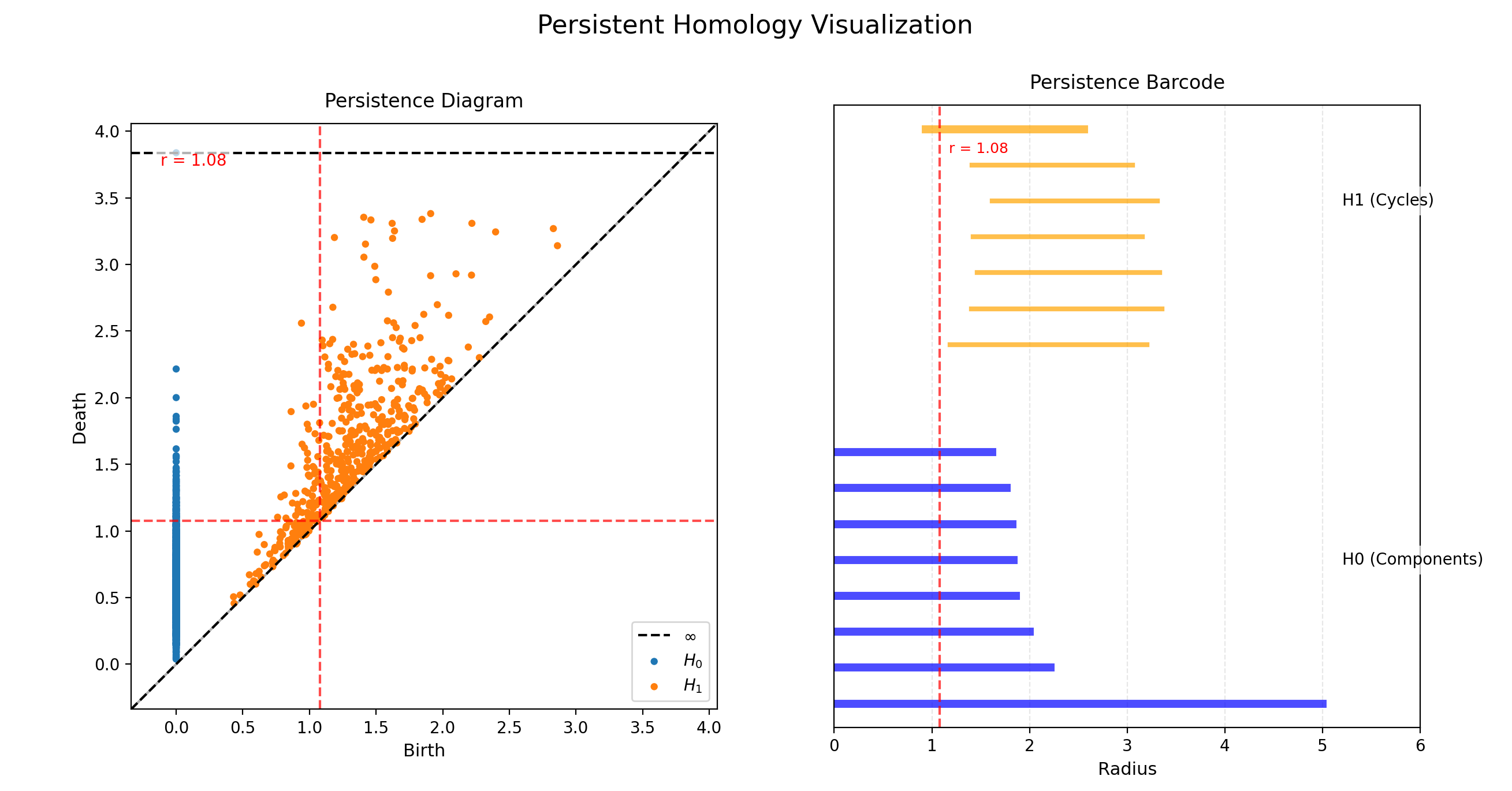}
    \caption{Persistence diagram and Persistence Barcodes generated from a Swiss roll dataset using \texttt{scikit-learn}~\cite{scikit-learn}. In the persistence diagrams, the blue points represent generators of $H_0$ cycles and the orange points represent generators of $H_1$ cycles, with  \(x\)-coordinates as birth radius and \(y\)-coordinates as the death radius. Features further from the diagonal are more persistent. In the persistence barcodes, the blue lines are generators of $H_0$ cycles and the orange lines are generators of $H_1$ cycles. The beginning of the line on the left represents the radius at the time of birth, and the end of the line on the right represents the radius at the time of death.}
  \label{fig:barcode}
\end{figure}

A larger persistence value indicates that a feature remains present across a wider range of the filtration and is thus more likely to be topologically meaningful. Conversely, features with birth and death times that occur in close succession ($pers_i \approx 0$) are more likely to be topological noise. To visually represent birth and death data, a barcode plot or a scatter plot can be used (see Figure~\ref{fig:barcode}). In both figures, each cycle is represented as its own data point. In the scatterplot, the x and y coordinates provide the birth and death times, respectively. In the barcode plot, the beginning of the bar gives the birth time and the end of the bar gives the death time. 

On top of using normal simplicial complexes, cubical complexes can be used for extracting image data to create a more natural representation. \cite{ziou_generating_2002}

Such a complex is built from elementary cubes of each dimension: points (0-cubes), edges/lines (1-cubes), squares (2-cubes), and various analogs for higher dimensions. As our images are 2-dimensional and grayscale, each pixel will correspond to a 2-cube. Its intensity value defines a scalar function over the image domain. Increasing the filtration parameter gradually includes pixels of greater intensity, thus growing the complex.

\begin{figure}[h]
  \centering
  \includegraphics[width=0.45\textwidth]{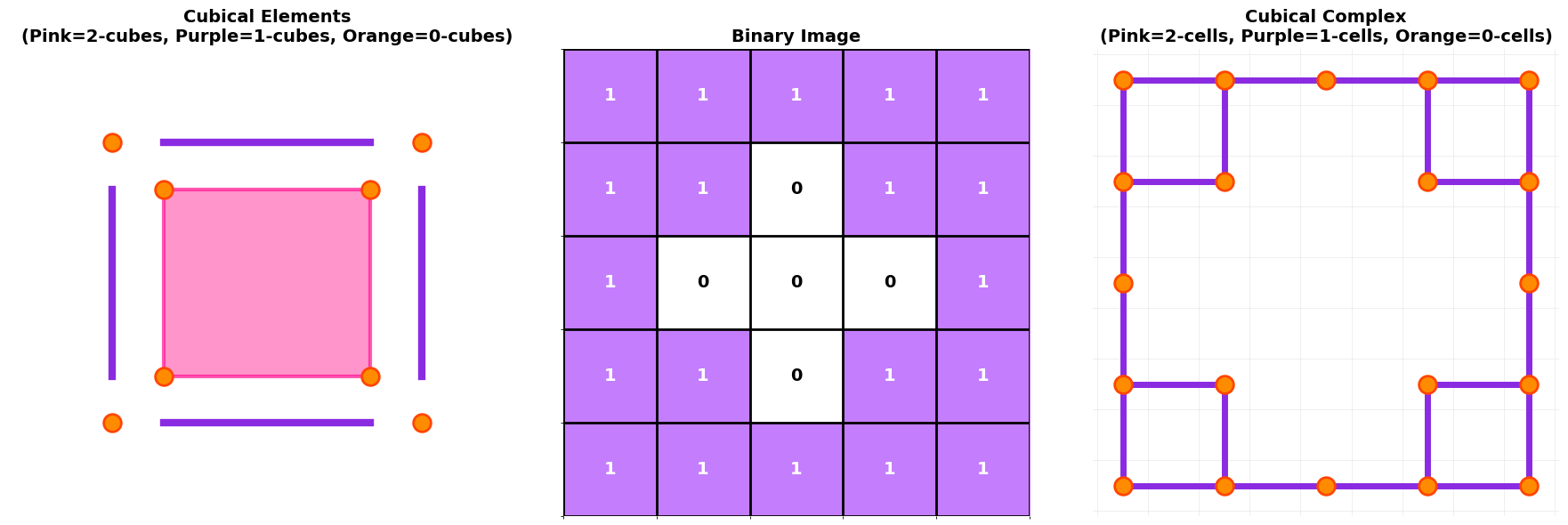}
    \caption{Binary image highlights foreground (value 1) and background (value 0) pixels. The cubical complex is built by placing cells around the foreground region, enabling topological analysis of the image’s structure through persistent homology. \cite{dongjin_figure_2023}}
  \label{fig:cubical_complex}
\end{figure}

Similar to simplicial complexes, we continue to track the topological features introduced in Subsection~\ref{PH}, specifically their births and deaths across the filtration. These events are recorded in a persistence diagram, which captures how the underlying shape and structure of the image evolve with changes in spatial resolution (see Figure~\ref{fig:cubical_complex}). This approach lends to an image's discrete, gridded structure, giving a computationally efficient manner by which we can compute the persistent homology.

\subsection{Wasserstein Distance}

Suppose we have two persistence diagrams $D_1 = \{x_1, x_2, ..., x_n\} \subseteq \mathbb{R}^2$ and $D_2 = \{y_1, y_2, ..., y_n\} \subseteq \mathbb{R}^2$, where $x_i$ and $y_i$ are points given by $(b_i, d_i)$. Here, $b_i$ is the birth time of the $i$-th feature, and $d_i$ is its death time.

The Wasserstein Distance allows us to measure the difference between two different persistence diagrams \cite{Berwald_Gottlieb_Munch_2018}, thus quantifying their topological similarity, which is given by:
\begin{center}
$W_p(D_1, D_2) = \left( \inf_{\gamma: D_1 \to D_2} \sum_{x \in D_1} \| x - \gamma(x) \|^p \right)^{1/p}$
\end{center}
Where $\gamma$ ranges over all bijections between $D_1$ and $D_2$. This cost function, which is the p-norm difference inside the Summation, evaluates the total effort required to morph one diagram into the other, measured by summing the distances between matched features. The $p$-th power in the summation allows the metric to adjust sensitivity to outliers or large deviations in feature persistence.

Regardless of the cardinalities of the diagrams, the optimal matching may pair some points to the diagonal $\Delta = \{(x,x) \in \mathbb{R}^2\}$, which represents features that are born and die at the same moment. Matching a point to $\Delta$ treats that feature as noise rather than forcing it to match an unrelated point in the other diagram. In practice, this is done by giving the diagonal infinite multiplicity, ensuring that the optimization can discard such occurrences if doing so reduces the total transportation cost.

The goal of minimizing the total cost over all bijections ensures that the Wasserstein distance reflects the optimal alignment of persistent features. Smaller values of $W_p$ indicate that the diagrams are topologically similar, while larger values reveal substantial structural differences in the underlying data. We used the $1$-Wasserstein metric ($p = 1$) because its linear treatment of distances reduces the influence of outliers, making comparisons of diagrams with short-lived or noisy features more stable and interpretable.

\subsection{Fourier Transform}
To identify the most significant frequencies, we apply the \emph{Fourier Transform}, which converts data from the time domain to the frequency domain and reveals the dominant frequency components useful for image feature extraction.

The Fourier Transform of a function $f: \mathbb{R}^2\rightarrow \mathbb{R}$ is given by:
$$
{F}\{f(t)\} = F(\omega) = \int_{-\infty}^{\infty} f(t) e^{-i \omega t} dt
$$
where $\boldsymbol{\omega} \in \mathbb{R}^2$ denotes the angular frequency vector, because we are working with 2D images. The Fourier Transform is a function 
$\mathcal{F} : L^1(\mathbb{R}) \rightarrow \mathbb{C},$
mapping an integrable function $f(t)$ to a complex-valued function $F(\omega)$ that encodes the amplitude and phase of its frequencies.

In practical applications, data is typically sampled at discrete intervals. The \emph{Discrete Fourier Transform} (DFT),  a discretized variant of the continuous Fourier Transform returns a set of numbers, called Fourier coefficients, rather than a function, enabling frequency analysis of digital signals.

The DFT is defined as:
\begin{equation} \label{eq:dft-def}
    X[k] = \sum_{n=0}^{N-1} x[n]  e^{\frac{-2\pi jkn}{N}}, \quad k = 0, \ldots, N-1
\end{equation}

where:
\begin{itemize}
    \item $x[n]$ is the original time-domain signal at index $n$,
    \item $N$ is the total number of samples,
    \item $X[k]$ is the amplitude and phase of the $k$th frequency component.
\end{itemize}

\begin{figure}[h]
  \centering
  \includegraphics[width=0.45\textwidth]{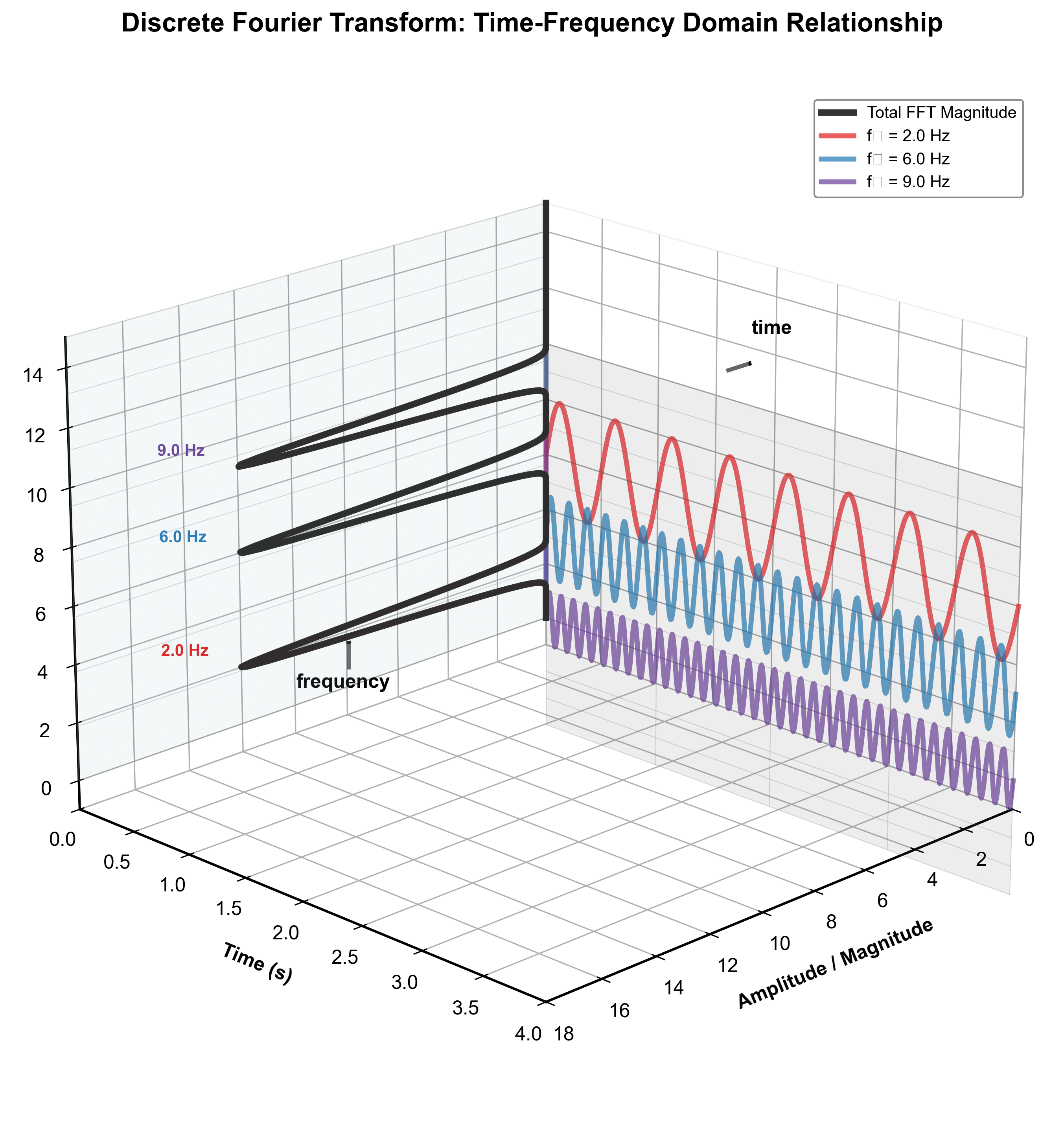}
  \includegraphics[width=0.45\textwidth]{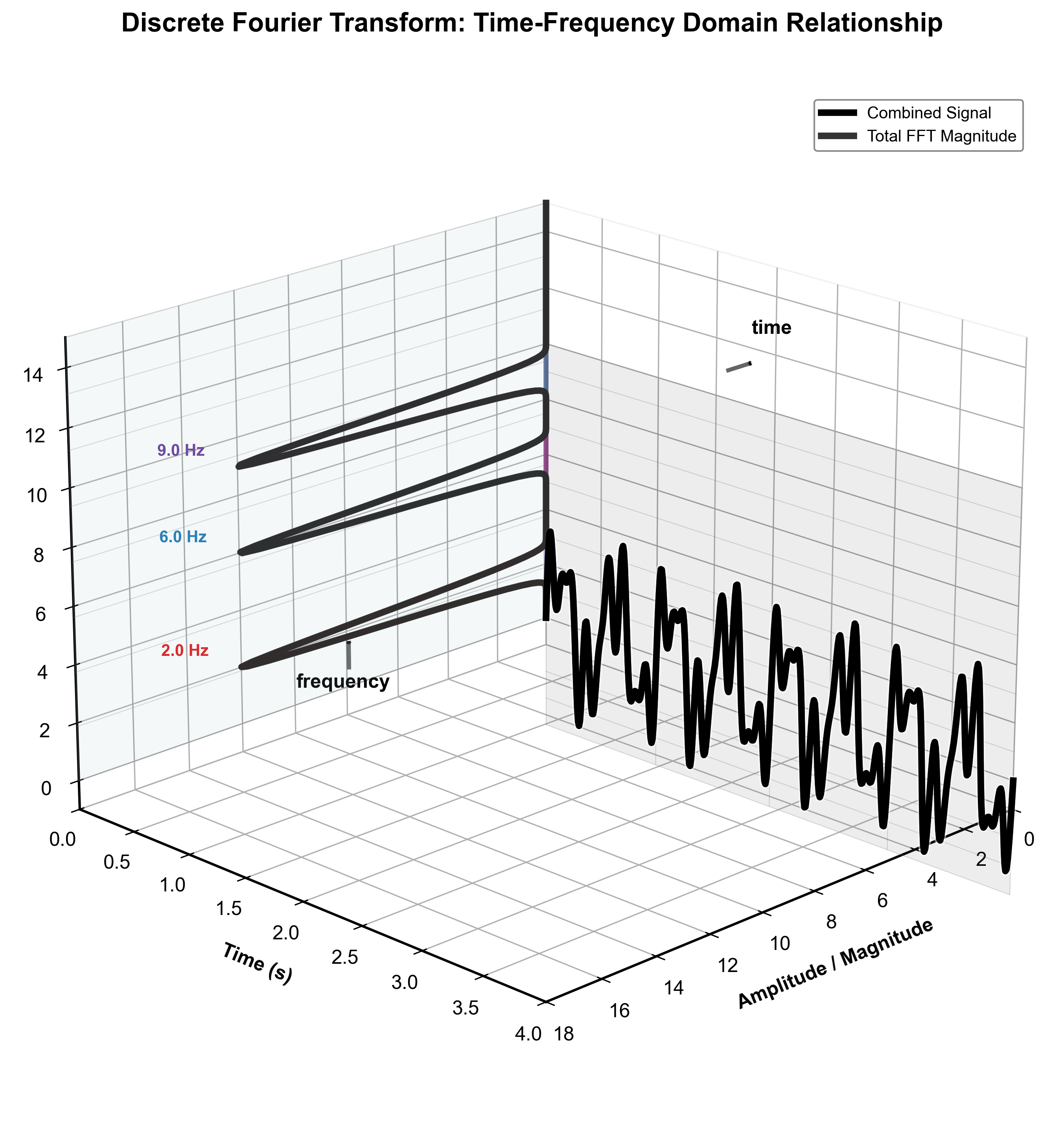}
  \caption{Diagram of the FFT and how it computes frequency from time where the colored lines represent different input frequencies and the black line represents the output frequency.}
  \label{fig:fft}
\end{figure}

The Discrete Fourier Transform (DFT) transforms a finite sequence of length \(n\) into another length-\(n\) sequence of complex numbers, representing the amplitude and phase of various frequency components present in the original signal. While the DFT has a time complexity of \(O(n^2)\), it is most often computed using the \textit{Fast Fourier Transform (FFT)} (see Figure~\ref{fig:fft}), which reduces the time complexity to \(O(n \log n)\) by exploiting symmetries and redundancies in the DFT's computations.

One key motivation behind the development of the FFT is the \emph{convolution theorem}, which states that convolution in the time domain of two functions corresponds to pointwise multiplication of the two functions' Fourier transform in the frequency domain. This theorem underpins many signal processing techniques and highlights the computational benefit of using the FFT: by transforming signals into the frequency domain, convolution operations—which are computationally expensive in the time domain—can be performed much more efficiently. 

\textbf{Convolution Theorem.}
Let \( x[n] \) and \( h[n] \) be two discrete-time signals, and let \( X[k] \) and \( H[k] \) be their respective DFTs. Then the DFT of their circular convolution is the pointwise product of their DFTs:
\begin{equation}
\text{DFT}(x[n] * h[n]) = X[k] \cdot H[k]
\end{equation}

\noindent
This means that convolution in the time domain becomes multiplication in the frequency domain, enabling faster computation using the FFT followed by an inverse FFT \cite{herman_convolution}.

The most widely used FFT algorithm, the \textit{Cooley--Tukey method}~\cite{cooley1965algorithm}, recursively divides a DFT of size \(N\) into smaller DFTs of size \(N/2\) by separating even and odd indexed elements. This divide-and-conquer approach continues until reaching DFTs of size 2, which are trivial to compute. The FFT ultimately decomposes the time-domain signal into a sum of sinusoidal components, each with a fixed amplitude and phase.

Although the FFT computes the same result as the DFT, it is more efficient and is the preferred choice in most practical applications, especially when working with large datasets. The DFT, however, remains fundamental for theoretical analysis and for contexts where the overhead of FFT setup (restructuring or padding in order to fit the FFT's structure) is not justified due to computation time, such as with very short or irregularly sampled signals. While the DFT defines the transformation, the FFT is the algorithmic breakthrough that enables its widespread real-time use.

Table \ref{fts_comparison} provides an overview of the discussed Fourier Transformations:

\begin{table}[h!]
\centering
\caption{Comparison of Fourier Transform Types}
\label{fts_comparison}
\scriptsize
\renewcommand{\arraystretch}{1.5} 
\begin{tabular}{|p{2.5cm}|p{3cm}|p{4cm}|p{3cm}|}
\hline
\textbf{Type} & \textbf{Fourier Transform} & \textbf{Discrete Fourier Transform} & \textbf{Fast Fourier Transform} \\
\hline
\textbf{Time Complexity} & N/A & $O(n^2)$ & $O(n \log n)$ \\
\hline
\textbf{Uses} & Continuous signals & Digital signal and image processing & Fast DFT computation \\
\hline
\textbf{Equation} & $F(\omega)=\int f(t)e^{-i\omega t} dt$ & $c_k=\frac{1}{N}\sum f_j e^{-i2\pi jk/N}$ & Same as DFT (optimized) \\
\hline
\end{tabular}
\end{table}

\subsection{Metrics for Quality Comparison}
\label{sec:Metrics for Quality Comparison}
We must define metrics to compare the similarity of the compressed image to that of the original image. This allows us to determine the trade off between the compression amount (by percentage of frequencies maintained) and the quality of the resulting image. 
\begin{enumerate}
\item \textbf{Structural Similarity (SSIM) index}\\
The SSIM index is perceptual metric which combines data from luminance, contrast, and structural comparisons. For image patches $x$ and $y$, which are $11\times11$ groups of pixels, the SSIM is defined as 
\[
\text{SSIM}(x, y) = \frac{(2 \mu_x \mu_y + C_1)(2 \sigma_{xy} + C_2)}{(\mu_x^2 + \mu_y^2 + C_1)(\sigma_x^2 + \sigma_y^2 + C_2)}
\]
where:
\begin{itemize}
\item $\mu_x, \mu_y$: The local means (averages) of the pixel intensities in image patches $x$ and $y$.
\item $\sigma_x^2, \sigma_y^2$: The local variances of the pixel intensities in image patches $x$ and $y$.
\item $\sigma_{xy}$: The covariance between the two patches, $x$ and $y$.
\item $C_1, C_2$: Small, fixed stabilizing constants used to prevent division by zero.
\end{itemize}
SSIM is a strong heuristic as it produces scores between $[0,1]$, where a score of $1$ indicates an identical picture. This metric is better used as a way to compare how humans see pictures, making it a good way to see how effective compression is at keeping the picture similar.
\item \textbf{Mean Squared Error (MSE)}\\
Mean squared error is a numerical metric which looks at the pixel-level differences. For two images $I_1$ and $I_2$ of size $N$, it is defined as 
\[
\text{MSE}(I_1, I_2) = \frac{1}{N} \sum_{i=1}^{N} \left( I_1[i] - I_2[i] \right)^2
\]
where:
\begin{itemize}
\item $N$: The total number of pixels in the image.
\item $I_1[i], I_2[i]$: The intensity (pixel) value of the $i$-th pixel in image $I_1$ and $I_2$, respectively.
\end{itemize}
MSE calculates the average squared deviation between the two images, with a smaller value indicating greater similarity. This metric is simple, but is sensitive to small deviations that can be caused by noise or illumination changes, which makes it not as effective for images where structure is more relevant than the actual pixel alignment.
\item \textbf{Bottleneck Distance}\\
The Bottleneck Distance is a topological metric derived from persistent homology as shown earlier which uses the birth and death points in a persistence diagram. Bottleneck Distance is calculated by measuring the maximum distance that any feature of the new image must be shifted in order to optimally match the features of the original image. In order to calculate this score when given an image $I$, we can first define a filtration of the sublevel sets such that $I_\alpha = \{ x \mid I(x) \le \alpha \}$ for varying thresholds $\alpha$. From this, after letting $b$ represent birth and $d$ represent death the bottleneck distance between diagrams $D(I_1)$ and $D(I_2)$ is defined as: 
\[
d_B(D(I_1), D(I_2)) = \inf_{\gamma} \sup_{x \in D(I_1)} \| x - \gamma(x) \|_\infty
\]
where:
\begin{itemize}
\item $D(I_1), D(I_2)$: The persistence diagrams for image $I_1$ and $I_2$.
\item $x$: An individual persistence feature (a birth-death point) in $D(I_1)$.
\item $\gamma$: A bijection (a one-to-one matching) between the points in $D(I_1)$ and $D(I_2)$.
\item $\inf_{\gamma}$: The infimum (greatest lower bound) over all possible matchings $\gamma$.
\item $\sup_{x}$: The supremum (least upper bound) of the distance for a single matched pair.
\item $\|\cdot\|_\infty$: The $L_\infty$ (maximum) norm, which measures the largest difference between birth or death times.
\end{itemize}
This metric allows us to focus on big differences within the image rather than small noise. 
\item \textbf{Betti Number Distance}\\
The Betti Number Distance is another topological metric which compares the images using their Betti numbers, which count the number of $k$-dimensional topological features. For a filtration of an image $I_\alpha$, the Betti numbers can be computed as functions $\beta_k(\alpha)$. After defining the two images $I_1$ and $I_2$, the Betti Number Distance is defined as the $L_p$ norm of the difference of these functions: 
\[
d_{\beta_k}(I_1, I_2) = \left( \int_\alpha \big| \beta_k^{I_1}(\alpha) - \beta_k^{I_2}(\alpha) \big|^p \, d\alpha \right)^{1/p}
\]
where:
\begin{itemize}
\item $\beta_k^{I_1}(\alpha), \beta_k^{I_2}(\alpha)$: The Betti number of dimension $k$ for image $I_1$ and $I_2$ at filtration value (threshold) $\alpha$.
\item $\int_\alpha$: An integral over all possible filtration values (thresholds) $\alpha$.
\item $p$: The exponent defining the specific $L_p$ norm (e.g., $p=1$ for $L_1$ distance).
\end{itemize}
This metric is good at showing how important topological features based on scale, which gives an overall score which is able to take into account larger picture changes instead of small intensity/quality changes.
\end{enumerate}

\section{Methodology}
\subsection{Image Preprocessing}
To ensure homology-based analysis is both computationally feasible and topologically meaningful, images must be properly preprocessed. In our study, the first step is converting the input image from RGB to grayscale. RGB images contain three color channels—red, green, and blue—each of which adds to the data complexity. Grayscale conversion reduces this to a single intensity channel, simplifying the data while preserving structural features necessary for topological analysis.

This transformation retains luminance information and removes color, enabling more efficient computation of the cubic complex and the Fourier transform. By working with a single-channel representation, we reduce computational overhead and make it easier to extract meaningful topological features from the image.

The images are also taken from 3 datasets and sorted by random to try to capture a larger range of images used in testing. We resize them all to $128$x$128$ to ensure a consistent number of frequencies is being checked. The datasets are Tiny-ImageNet \cite{tinyimagenet2023}, CIFAR-10 \cite{cifar10}, and STL-10 \cite{coates2011stl10}.

\subsection{Persistence Diagrams}
Next, for each image the cubical complex is generated by our program to give a full persistence diagram. This is a relatively quick, but critical step. Then, the FFT is calculated on the grayscaled image and stored. These two computations stored will be very important in the next step.

\subsection{Ranking Frequencies}
To reconstruct the image efficiently, we need to select the most important frequencies.
We begin by computing all frequency indices of the FFT spectrum.
For each frequency pair $(f_x, f_y)$, we reconstruct the image using only that frequency and its conjugate, then compute its persistence diagram. We refer to $(f_x, f_y)$ as a pair because a 2-D Fourier transform indexes each mode by its horizontal and vertical frequencies, which together define a single two-dimensional sinusoidal component.

We then calculate the Wasserstein distance $W_1(D_{\text{full}}, D_{\text{single}})$ between the full persistence diagram $D_{\text{full}}$ and the diagram $D_{\text{single}}$ of the single frequency, which measures how much that individual frequency contributes to the overall topological structure of the image.

The importance score of each frequency is computed as:
\begin{equation}
    score = W_1(D_{\text{full}}, D_{\text{single}}) \cdot \frac{1}{\sqrt{f_x^2 + f_y^2}}
\end{equation}
Here, the first term $W_1$ provides a topological measure of importance, selecting frequencies that encode meaningful structural features (loops and components). The second term $\frac{1}{\sqrt{f_x^2 + f_y^2}}$ gives higher weight to low-frequency components, serving as a structural prior that aligns the score with the known $1/f$ spectral energy distribution of natural images. This design ensures that the most perceptually relevant information, which resides primarily in the low-frequency domain, is inherently favored by the ranking, preventing high-frequency noise from dominating the selection.

Finally, all frequencies are sorted by this score in descending order, and the highest-scoring frequencies are retained for reconstruction.
The results from these approaches are discussed in~\ref{results}.

\subsection{Image Compression}
This method takes inspiration from JPEG image compression. JPEG's process involves two key stages: 1) lossy filtering via the Discrete Cosine Transform (DCT), which is used to transform image blocks into the frequency domain and remove high-frequency coefficients, and 2) a final lossless compression step using Huffman coding (entropy coding). The DCT, being similar to the DFT, is used to eliminate high-frequency information that is less perceptible to the human eye \cite{AlAniimagecompression}. This lossy frequency-domain filtering step is analogous to our frequency ranking, but critically, the subsequent Huffman coding exploits statistical redundancy in the remaining data to drastically reduce the final file size, a feature currently absent in our raw frequency retention method. This explains the significant disparity in file size results compared to JPEG. Additionally, low frequencies are more prevalent in natural images (those taken from the real world), because there is less contrast in pixel intensity between adjacent pixels. This is an application of low pass filtering. 

In some cases, this is not the best way to approach image compression. In digitally generated images, the contrast from pixel-to-pixel cannot be generalized as easily as in natural images. If one needs to compress images more accurately (that is, compression with respect to the unique topological properties of the image), we propose \textbf{homology-based compression} (PH Compression).

The compression is based on keeping a percentage of frequencies and then comparing it to JPEG compression using a parameter quality equal to the percentage of frequencies kept (though the quality is clamped between 5 and 95). This gives a better comparison against traditional methods of compressing images, such as JPEGS. 

A Gaussian smoothing corresponds to convolving the image with a Gaussian kernel
\[
G_\sigma(x,y) = \frac{1}{2 \pi \sigma^2} e^{(-\frac{x^2+y^2}{2\sigma^2})},
\]
where $\sigma$ controls the standard deviation (width) of the smoothing. This process reduces high-frequency noise while preserving the global structure of the image, stabilizing the persistence diagram. To ensure persistence diagrams are not dominated by high-frequencies during compression, a light Gaussian smoothing ($\sigma = 1 $) is applied to the PH-based reconstruction before computing its persistence diagram. To ensure persistence diagrams are not dominated by high-frequency noise during compression, a light Gaussian smoothing ($\sigma = 1$) is applied to the PH-based reconstruction before computing its persistence diagram. 
This stabilizes topological features and reduces noise without significantly altering the global structure of the image for each percentage.

While computationally expensive, we hope that this homology-based processing will allow for more precise lossy compression than JPEG (simply removing high frequencies). For a better understanding of our process, see Algorithm ~\ref{alg:ph_compression}

\begin{algorithm}[h]
\caption{Homology-Based Image Compression with FFT and Persistence Diagrams}
\label{alg:ph_compression}
\begin{algorithmic}[1]
\REQUIRE Image dataset $\mathcal{D}$, target compression ratio $\alpha$
\ENSURE Compressed image preserving topological features
\STATE \textbf{Preprocessing:}
\FOR{each image $I \in \mathcal{D}$}
    \STATE Convert $I$ from RGB to grayscale
    \STATE Resize $I$ to $128 \times 128$
\ENDFOR

\STATE \textbf{Persistence Diagram and FFT:}
\FOR{each image $I \in \mathcal{D}$}
    \STATE Compute cubical complex and full persistence diagram $PD_{full}$
    \STATE Compute 2D FFT of $I$: $F = \text{FFT}(I)$
\ENDFOR

\STATE \textbf{Frequency Ranking:}
\FOR{each frequency pair $(f_x,f_y)$ in $F$}
    \STATE Reconstruct image using only $(f_x,f_y)$ and its conjugate
    \STATE Compute persistence diagram $PD_{freq}$
    \STATE Compute Wasserstein distance: $W = W_1(PD_{full}, PD_{freq})$
    \STATE Compute importance score:
    \[
        score = W \cdot \frac{1}{\sqrt{f_x^2 + f_y^2}}
    \]
\ENDFOR
\STATE Sort frequencies by $score$ (ascending)

\STATE \textbf{Compression:}
\STATE Retain top $\alpha \cdot |F|$ frequencies based on score
\STATE Reconstruct compressed image from selected frequencies
\STATE Apply Gaussian smoothing with $\sigma = 1$
\STATE Compute persistence diagram of compressed image

\STATE \textbf{Comparison:}
\STATE Compare PH-based compression to JPEG at equivalent quality level
\end{algorithmic}
\end{algorithm}

\section{Code Implementation}
Our codebase, including implementations of persistent homology computation, and Wasserstein distance calculations, is publicly available on GitHub at \href{https://github.com/RMATH3/persistent-homology-compression.git}{https://github.com/RMATH3/persistent-homology-compression.git}. This repository includes scripts for image processing and compression. 
\subsection{Libraries and Tools}
Our implementation utilizes several key libraries and tools for computational topology and image processing. NumPy provides the foundation for numerical operations, including fast Fourier transforms (FFTs) and array manipulation. For visualization, we employ Matplotlib to display original, frequency-domain, and reconstructed images. Image preprocessing tasks such as reading, grayscaling, and resizing are handled through scikit-image. 

The core topological computations rely on GUDHI for building cubical complexes and computing persistent homology diagrams. Within the GUDHI framework, we use the Wasserstein distance functionality for comparing persistence diagrams and determining the significance of frequency components. Additionally, we incorporate standard Python libraries including OS for file existence checks and basic I/O operations, and time to record the duration of each compression and denoising operation.

\section{Results} \label{results}
\subsection{Results Overview}

\begin{figure}
    \centering
    \includegraphics[width=0.95\linewidth]{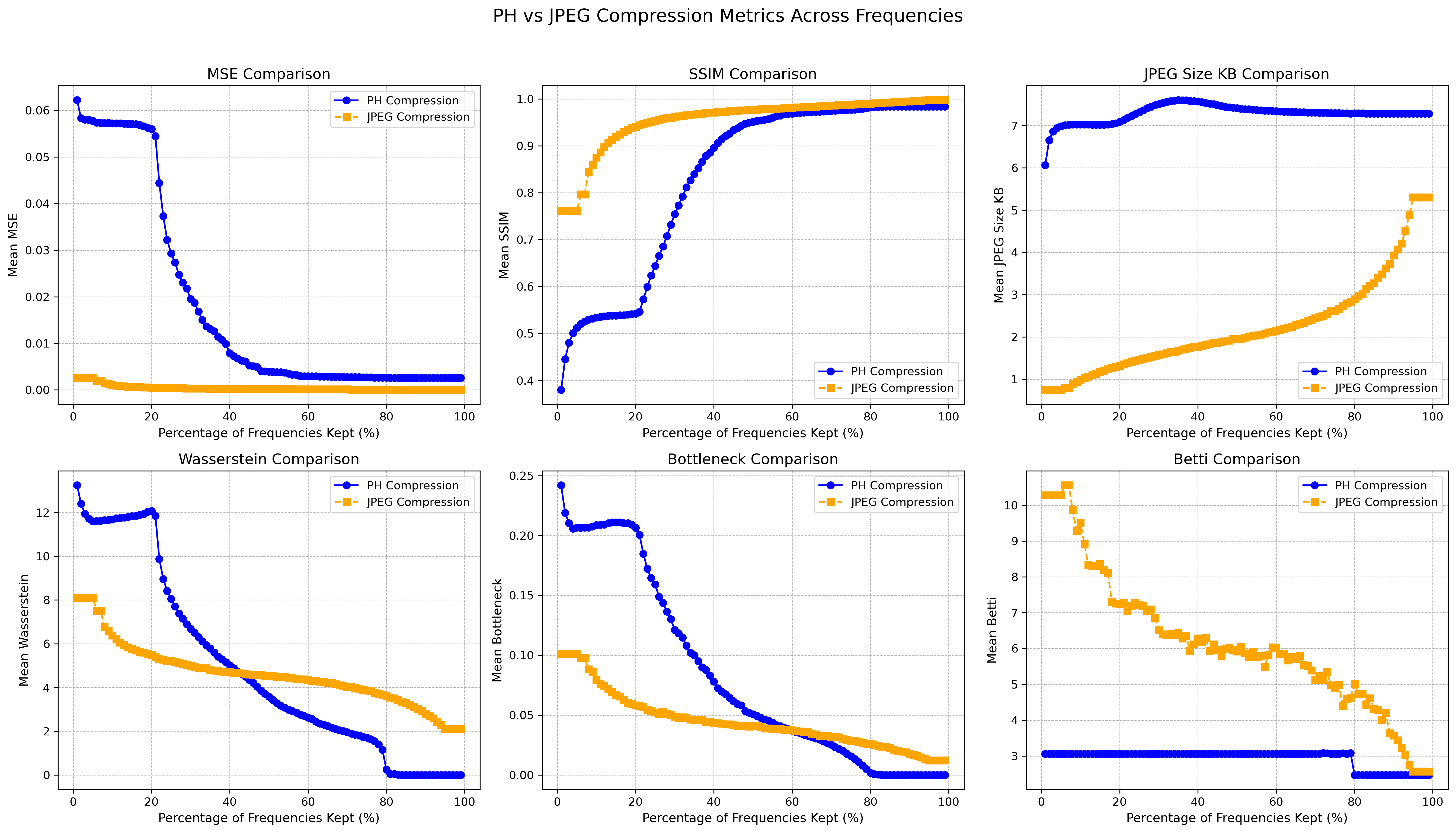}
    \caption{Comparison of JPEG and PH compression across 100 diverse images sampled from TinyImageNet, CIFAR-10, and STL-10. Results are evaluated using six metrics: four defined in Section~\ref{sec:Metrics for Quality Comparison}, Wasserstein Distance, and file size.}
    \label{fig:al_data}
\end{figure}

We compare JPEG compression and PH compression using six robust metrics: the four metrics defined in Section~\ref{sec:Metrics for Quality Comparison}, Wasserstein Distance, and file size. The analysis utilizes randomized training sets from TinyImageNet, CIFAR-10, and STL-10, providing a large pool of images for comparison. Due to the computational complexity of our process, which scales as $\mathcal{O}(n^3)$, we limit the evaluation to a subset of 100 images. Despite this limitation, we consider the sample sufficient to yield meaningful results, given the diversity of images included in the analysis. (See Fig. ~\ref{fig:al_data}).

\subsection{Visual Metrics}

Figure~\ref{fig:al_data} illustrates that, for approximately 30--50\% of retained frequencies, both the MSE and SSIM metrics for PH compression approach the values observed for JPEG compression. This outcome is expected, as JPEG primarily removes high-frequency components without significantly affecting visual quality. Beyond approximately 40\% retention, both the quantitative metrics and qualitative inspection of the images (see Section~\ref{sec:An example}) indicate minimal visual difference between PH compression and JPEG. These results suggest that PH compression maintains visual fidelity comparable to JPEG when a sufficiently high proportion of frequencies is preserved.

\subsection{Topological Metrics}

Analysis of the Wasserstein Distance, Bottleneck Distance, and Betti Number metrics reveals several notable trends. Both the Wasserstein and Bottleneck metrics show superior performance for PH compression relative to JPEG once approximately 50\% of frequencies are retained, and both metrics approach zero at 80\% retention. PH compression likely outperforms JPEG in these topological metrics because it prioritizes the preservation of persistent topological features rather than individual pixel values, so retaining a high proportion of significant frequencies directly benefits the topological structure.

\subsection{File Size}

The file size exhibits an interesting and interesting trend for PH compression. The size increases when the retained frequency \% is between 20\% to 35\%, before decreasing again at higher percents. In contrast, JPEG compression demonstrates a generally linear increase in file size with higher quality settings, except at above 80\% when it begins to grow exponentially. It might seem more favorable than PH compression from a strict file size perspective. However, when considering the broader quality metrics, this advantage is mitigated.

This trend in PH compression is likely a result of how spatial redundancy interacts with frequency retention. At very low retention levels, only a few dominant low-frequency components are preserved, producing smooth reconstructions with high spatial redundancy, which compress efficiently. As more frequencies are introduced around 20–35\%, the reconstruction gains details but remains far from fully structured since it introduces mid-frequencies. This introduces irregular patterns and localized variations, which reduce spatial redundancy and make the image harder to compress, leading to a temporary increase in file size. Beyond this range, as higher frequencies are incorporated, the reconstructed image begins to resemble the original image more closely, restoring spatial redundancy and enabling the lossless format to compress the data more effectively. This is why the file size decreases again at higher retention levels.

\subsection{An Example}
\label{sec:An example}

\begin{figure}[h]
    \centering
    \includegraphics[width=0.3\linewidth]{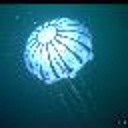}\hspace{0.05\linewidth}%
    \includegraphics[width=0.55\linewidth]{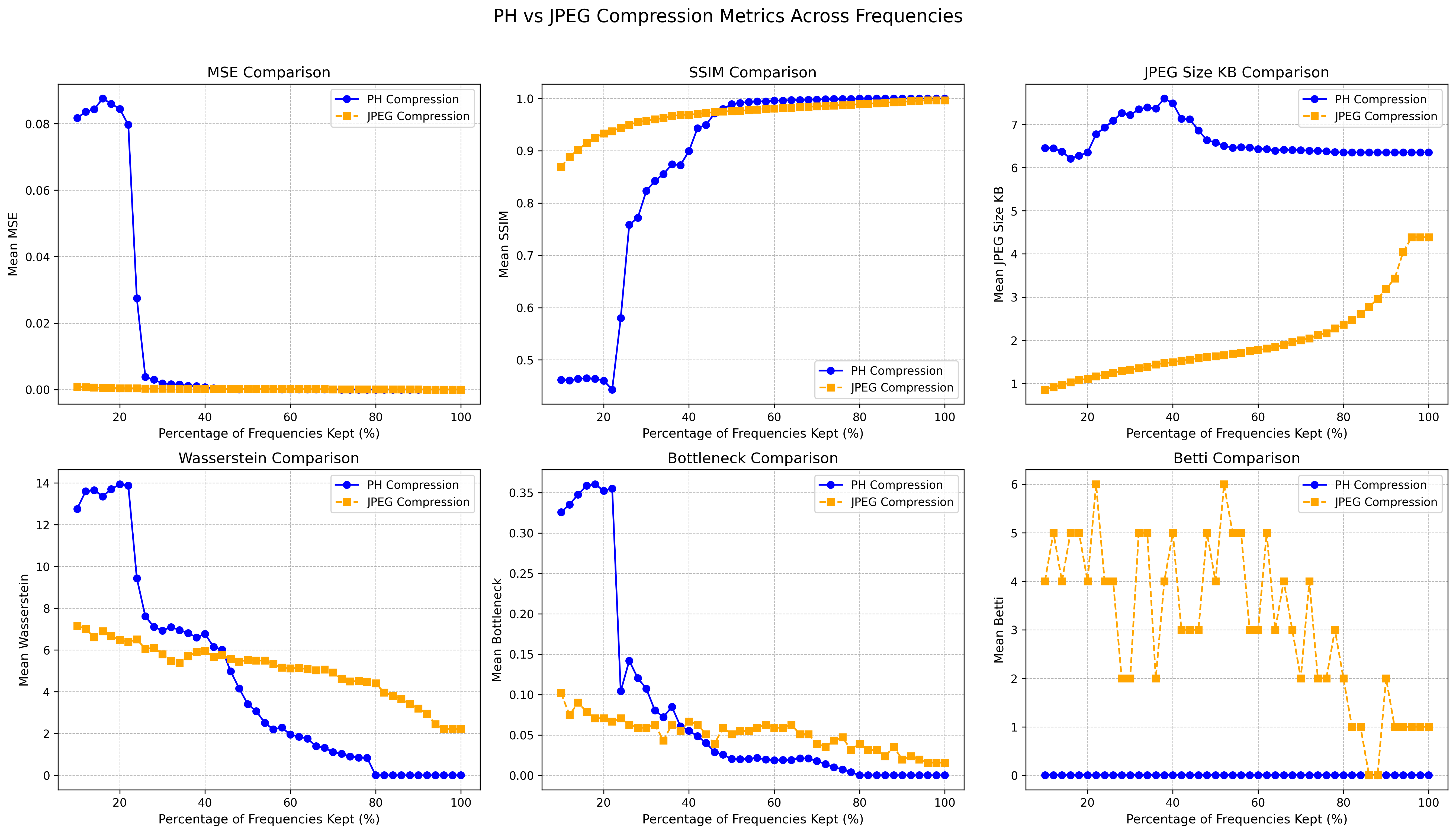}
    \caption{Left: Original image used, taken at random from the combination of the three datasets. Right: Comparison of all six metrics for the example jellyfish image.}
    \label{fig:Jellyfish_combined}
\end{figure}

\begin{figure}[h!]
\centering
\begin{minipage}{0.2\linewidth}
    \centering
    \includegraphics[width=\linewidth]{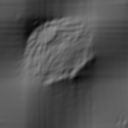}\\
    \small PH (20\%)
\end{minipage}
\begin{minipage}{0.2\linewidth}
    \centering
    \includegraphics[width=\linewidth]{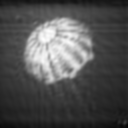}\\
    \small PH (30\%)
\end{minipage}
\begin{minipage}{0.2\linewidth}
    \centering
    \includegraphics[width=\linewidth]{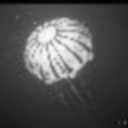}\\
    \small PH (90\%)
\end{minipage}

\vspace{0.5em}

\begin{minipage}{0.2\linewidth}
    \centering
    \includegraphics[width=\linewidth]{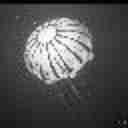}\\
    \small JPEG (20\%)
\end{minipage}
\begin{minipage}{0.2\linewidth}
    \centering
    \includegraphics[width=\linewidth]{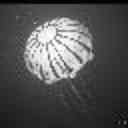}\\
    \small JPEG (30\%)
\end{minipage}
\begin{minipage}{0.2\linewidth}
    \centering
    \includegraphics[width=\linewidth]{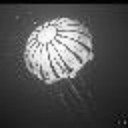}\\
    \small JPEG (90\%)
\end{minipage}
\caption{Comparison of reconstructions at different compression levels.}
\label{fig:ph_vs_jpeg}
\end{figure}

Using an image of a jellyfish from the dataset, we observe that PH compression outperforms JPEG compression both visually and topologically, as summarized in Fig.~\ref{fig:Jellyfish_combined}, Fig.~\ref{fig:ph_vs_jpeg}, and Table~\ref{tab:file_sizes}. The original jellyfish image is used as a reference. The results for this example follow the same trends observed across the larger dataset regarding the proportion of frequencies required for PH compression to match or exceed the performance of JPEG. It shows promising results, especially that visually, preserving frequencies over 25\% leads to images that are nearly identical to JPEGs.

\begin{table}[h]
\centering
\caption{File sizes (in KB) for the original image, PH compression, and JPEG compression at different compression levels.}
\begin{tabular}{lcccc}
\toprule
\textbf{Compression Type} & \textbf{Original} & \textbf{20\%} & \textbf{30\%} & \textbf{90\%} \\
\midrule
\textbf{Original} & 21 KB & -- & -- & -- \\
\textbf{Persistent Homology} & -- & 6.5 KB & 7.4 KB & 6.5 KB \\
\textbf{JPEG} & -- & 1.1 KB & 1.4 KB & 3.3 KB \\
\bottomrule
\end{tabular}

\label{tab:file_sizes}
\end{table}

\subsection{PH Compression vs JPEG Compression}

Considering all the metrics analyzed, PH compression demonstrates several notable advantages. The PH-compressed images are virtually indistinguishable from JPEG images visually, while exhibiting superior performance on topological metrics when a high proportion of frequencies is retained. Although the current file sizes for PH compression are larger, which may pose challenges for large-scale applications, the method shows significant potential for computer vision and machine learning tasks. This is particularly relevant in fields that prioritize topological information over purely visual fidelity, such as healthcare and scientific imaging.

\section{Future Work}
This study introduces a framework for frequency-domain image filtering guided by persistent homology. Several promising directions remain for future exploration:

\textbf{Topological Feature-Based Classification:} 
    Although homology-guided filtered images may potentially lose some original detail visually, their noise reduction could improve classification performance and should be tested. Unlike JPEG compression, which is optimized for human perception, persistent homology filtering might be better suited for computer analysis.

\textbf{Real-Time or Low-Power Implementations:} Given the computational intensity of Wasserstein-based filtering, future work may explore real-time or embedded implementations using approximate distance metrics or hardware acceleration.

\textbf{Integration with Other Topological Tools:} Additional topological descriptors such as Betti curves, persistence landscapes, or mapper graphs could complement persistent homology to improve preprocessing, especially in higher-dimensional or multi-channel data.

\textbf{Denoising} 
    The combination of persistent homology and the Fourier transform can also be applied to denoising. Instead of traditional denoising methods, a topologically-guided approach avoids blanket thresholds on pixel values and the trial-and-error involved in handcrafting pixel values for each image. If one requires a denoising strategy tailored to an image's structure, we propose \textbf{homology-based denoising}. 

    Given that noise is not relevant for the underlying structural topology of an image, a homology-based filter could precisely remove the frequencies that correspond with noise, maintaining the frequencies that determine the actual content of the image.

\textbf{Huffman Coding}
    While persistent homology effectively reduces image size by discarding low-persistence (noise-like) frequencies, the resulting frequency-filtered representation can still contain statistically redundant information. Because the retained frequencies follow a highly non-uniform distribution—persistent features occupy only a small subset of the Fourier domain—further lossless compression is possible.

    Huffman coding, or related entropy-based methods, can be applied to the filtered frequency coefficients to exploit this sparsity. After persistent homology removes insignificant frequencies, many coefficients become zero or near-zero, producing a symbol distribution with a strong skew that is ideal for variable-length encoding. Applying Huffman encoding to the remaining high-persistence coefficients would reduce storage without altering the topological structure maintained by the homology filter.

    In combination with homology-based frequency selection, Huffman coding enables a two-stage compression pipeline: 
    
    \begin{enumerate}
        \item Topological reduction removes structurally irrelevant frequencies.
        \item Entropy coding compresses the remaining structured information.
    \end{enumerate}
    This approach may yield competitive compression ratios while preserving the essential topological features that the method aims to protect.

\section{Acknowledgments}
We would like to express our sincere gratitude to the North Carolina School of Science and Mathematics for their continued support and for providing the environment that made this research possible. We are especially grateful to Dr.Hannah Schwartz for her mentorship, guidance, and invaluable contributions throughout the duration of this project. We also extend our appreciation to Dr. Bryan Stutzman and Mr. Reed Hubbard for their consistent support at every stage of this work.

\bibliographystyle{unsrt}
\bibliography{references} 

\end{document}